\documentclass[letterpaper]{article} 
\usepackage{aaai23}  
\usepackage{times}  
\usepackage{helvet}  
\usepackage{courier}  
\usepackage[hyphens]{url}  
\usepackage{graphicx} 
\usepackage{natbib}  
\usepackage{caption} 
\usepackage{algorithm}
\usepackage{algorithmic}
\usepackage{multirow}

\usepackage{newfloat}
\usepackage{listings}
\DeclareCaptionStyle{ruled}{labelfont=normalfont,labelsep=colon,strut=off} 
\floatstyle{ruled}
\newfloat{listing}{tb}{lst}{}
\floatname{listing}{Listing}
\title{H-AES: Towards Automated Essay Scoring for Hindi}

\author {
    Shubhankar Singh\textsuperscript{\rm 1},
    Anirudh Pupneja\textsuperscript{\rm 2},
    Shivaansh Mital\textsuperscript{\rm 3},
    Cheril Shah\textsuperscript{\rm 4},
    Manish Bawkar\textsuperscript{\rm 5},
    Lakshman Prasad Gupta\textsuperscript{\rm 6},
    Ajit Kumar\textsuperscript{\rm 3},
    Yaman Kumar\textsuperscript{\rm 3},
    Rushali Gupta\textsuperscript{\rm 7},
    Rajiv Ratn Shah\textsuperscript{\rm 3}
}
\affiliations {
    \textsuperscript{\rm 1} Manipal University Jaipur,\\
    \textsuperscript{\rm 2} BITS Pilani, K K Birla Goa Campus,\\
    \textsuperscript{\rm 3} Indraprastha Institute of Information Technology, Delhi,\\	
    \textsuperscript{\rm 4} Pune Institute of Computer Technology,\\
    \textsuperscript{\rm 5} Sardar Vallabhbhai National Institute of Technology, Surat,\\
    \textsuperscript{\rm 6} University of Allahabad,\\
    \textsuperscript{\rm 7} Banaras Hindu University\\
}

\begin{document}

\maketitle

\everypar{\looseness=-1}

\begin{abstract}

The use of Natural Language Processing (NLP) for Automated Essay Scoring (AES) has been well explored in the English language, with benchmark models exhibiting performance comparable to human scorers. However, AES in Hindi and other low-resource languages remains unexplored. In this study, we reproduce and compare state-of-the-art methods for AES in the Hindi domain. We employ classical feature-based Machine Learning (ML) and advanced end-to-end models, including LSTM Networks and Fine-Tuned Transformer Architecture, in our approach and derive results comparable to those in the English language domain. Hindi being a low-resource language, lacks a dedicated essay-scoring corpus. We train and evaluate our models using translated English essays and empirically measure their performance on our own small-scale, real-world Hindi corpus. We follow this up with an in-depth analysis discussing prompt-specific behavior of different language models implemented.
\end{abstract}

\section{Introduction}
Academic assessments have long used short-text response and essay writing tasks, which have resulted in numerous scoring approaches, prompt design, and assessment methods. These tasks help judge many aspects of a student’s language learning abilities and are commonly integrated with curricula and standardized tests worldwide in multiple languages. Automated essay scoring (AES) emulates human judgment when evaluating the quality of these written essays. Traditional essay scoring methods entail a vast corpus of written data that human scorers manually evaluate. Manual scoring is a laborious task which is challenging to scale much beyond the limited classroom settings \cite{kumar2019get,zhang2013contrasting}. AES utilizes Natural Language Processing (NLP) and Machine Learning (ML) techniques to evaluate these essays in a more efficient and scalable manner. Commonly used in standardized tests like the GRE and the TOEFL \cite{Attali_Burstein_2006}, many organizations and education councils have turned to use AES to reduce workload \cite{singla2022using}.

Commonly, AES is facilitated by a vast set of training data which are scored using expert-designed evaluation rubrics \cite{weinberger2011analytical}. The scores assigned can be broadly categorized into two types: \textit{Holistic scores} - a single discrete score value from a given whole number range and \textit{Trait-based scores} - multiple score values assigned based on multi-dimensional criteria such as relevance to prompt, argument quality, and coherence \cite{ke2019automated,bamdev2022automated}. A majority of work done on the task has entailed holistic scoring \cite{ke2019automated}, leveraging the Automated Student Assessment Prize (ASAP) corpus\footnote{\url{https://www.kaggle.com/c/asap-aes}}, released on Kaggle in 2012. Eight essay prompts are included in the ASAP corpus covering a broad range of topics. Over time, it has become a widely used corpus for holistic scoring. Most approaches to the AES task are specific to each prompt. 

Although short-text response writing is a standard assessment task for students worldwide in different languages, most research has focused on the English language domain, with relatively few focusing on other languages and multilingual approaches. Research in AES for Indic languages is negligible. Hindi is the most spoken language in India, with around 528 million native speakers, according to the 2011 language census \cite{india2011census}. With millions in the country writing their school-level, graduation, and public-sector examinations in Hindi, there is a dire need to automate the scoring processes. In addition, with the rapidly expanding middle class in India, the number of Hindi telecallers is also increasing at an equally rapid pace. New-age unicorn startups like Apna which hire blue and white collar telecallers for the South Asian market, have instituted automatic scoring as the first filtering step. On the other hand, Hindi NLP methods for automatic scoring are still in the early development stages. The Hindi NLP space’s predicament is a veritable lack of data resources. An approach to address this issue is to use English data translated to Hindi, which has been commonly used to train large language models  \cite{kakwani2020indicnlpsuite,conneau-etal-2020-unsupervised,khanuja2021muril}. The need for feature engineering is removed by neural approaches which discover both simple and complex features on their own.


Since a dedicated corpus for AES in Hindi currently does not exist, we use a machine-translated version of the ASAP corpus to train, validate and test our approaches. Our study implements recent approaches used for English AES to the Hindi language domain. Further, to validate that despite translation, models trained on the translated data perform adequately in the natural Hindi language settings, we collect a small-scale natural Hindi corpus and test all our models on this corpus. We employ both classical feature-based models and advanced end-to-end language models based on LSTMs and transformer architectures. 

The following three sections explain the related material, the corpus used, and the methodologies employed to perform the task on different models. In Section-5, we explain our experimental procedures, provide empirical results and comparisons with relevant English AES benchmark results, and follow it up with a detailed discussion. Section-6 concludes this study and discusses the limitations and future directions that can be taken to build more accurate and robust AES systems in Hindi and other languages. 
The dataset, rubric and the source code are made publicly
available\footnote{\url{https://github.com/midas-research/hindi-aes}}.

\section{Related Work}
Research for AES in English has spanned decades \cite{shermis2013handbook,zupanc2016advances,ke2019automated}. Many studies have treated AES as a regression and text classification problem, while a few apply ranking-based approaches \cite{bamdevclassification,bamdev2022automated}. 

Classical machine learning techniques like linear regression \cite{miltsakaki2004evaluation} support vector regression 
\cite{persing2010modeling,persing2015modeling,cozma2018automated}, and sequential minimal optimization (SMO) \cite{vajjala2018automated} are used typically for regression-based AES. SMO \cite{vajjala2018automated}, logistic regression \cite{nguyen2018argument} and Bayesian network classification \cite{rudner2002automated} utilize classification approaches. For ranking, SVM ranking \cite{yannakoudakis2011new} and LambdaMART \cite{chen2013automated} have been used. These approaches use custom linguistic features such as errors in grammar \cite{Attali_Burstein_2006}, readability features \cite{zesch2015task}, length and syntactic features etc. These techniques are widely popular and have been used in the production of evaluation systems, like E-Rater \cite{Attali_Burstein_2006} used in high-stake examinations like GRE and TOEFL.

Progress in Deep Neural Networks, like Convolutional Neural Networks (CNNs), Recurrent Neural Networks (RNNs), and Long Short Term Memory networks (LSTMs), allow for better performance in AES systems \cite{taghipour2016neural,alikaniotis-etal-2016-automatic,dong-etal-2017-attention,wang2018automatic,tay2018skipflow,song2020hierarchical,mathias2020can}. RNNs and LSTMs are typically used for language processing due to their ability to process sequential data. LSTMs have demonstrated better performance than RNNs for longer sequences due to their ability to retain long-term dependencies. These methods do not require any handcrafted components, leading to their widespread adoption. In the case of end-to-end Deep Neural Networks, complex features are automatically discovered, saving the time spent on designing them manually. Out of these models, SKIPFLOW \cite{tay2018skipflow} is able to compete with state-of-the-art by capturing coherence, flow, and semantic relatedness over time, which the authors call neural coherence features. 

Lately, pre-trained language models, built using transformer architecture, such as GPT, BERT \cite{devlin2018bert}, and XLNet \cite{yang2019xlnet} have pushed forward the limits of language understanding and generation attributed to their excellent generalization and representation abilities. These models have achieved better results in text classification and regression tasks. Although initial studies using transformers for AES \cite{rodriguez2019language,uto2020neural,mayfield2020should} fail to demonstrate a significant advantage over other deep learning approaches, work from \cite{10.1145/3397271.3401037,yang2020enhancing,wang-etal-2022-use} leverage such pre-trained language models, particularly BERT, and have given state-of-the-art results outperforming various traditional and deep-learning architectures designed for AES. \citeauthor{10.1145/3397271.3401037} \shortcite{10.1145/3397271.3401037} proposes two self-supervised tasks and a domain adversarial training technique for training optimization. This is the first work that uses pre-trained language models to outperform LSTM-based methods significantly. R\textsuperscript{2}BERT \cite{yang2020enhancing} combines regression and ranking to fine-tune BERT and obtain the new state-of-the-art. \citeauthor{wang-etal-2022-use} \shortcite{wang-etal-2022-use} developed a novel multi-scale essay representation approach based on BERT, employing multiple losses and transfer learning. 

Attempts have been made in the past to develop AES for languages other than English. These include research in AES for Chinese \cite{schultz2013intellimetric,song2020multi}, Arabic \cite{alghamdi2014hybrid,azmi2019aaee}, Japanese \cite{hirao2020automated}, Swedish \cite{ostling-etal-2013-automated,smolentzov2013automated}, German \cite{zesch2015task,ludwig2021automated}, Portuguese \cite{amorim2017multi,fonseca2018automatically} and more. The lack of dedicated essay corpora as comprehensive as the ones in existence for  English is a common feature among most of these studies. While few studies mentioned have explicitly created their organic corpora to work on, many have devised alternate solutions such as scraping the internet for essays, distilling essays from articles and combining datasets used for other tasks. 

The progress made in language processing for Indic Languages is gradual but on course. A few notable developments include \citeauthor{bhattacharyya-2010-indowordnet} \shortcite{bhattacharyya-2010-indowordnet}, \citeauthor{arora2020inltk} \shortcite{arora2020inltk}, \citeauthor{kakwani2020indicnlpsuite} \shortcite{kakwani2020indicnlpsuite},
\citeauthor{ramesh2021samanantar} \shortcite{ramesh2021samanantar}, and more. \citeauthor{DBLP:journals/corr/abs-2102-00214} \shortcite{DBLP:journals/corr/abs-2102-00214} presents a comprehensive report on the advancements in Hindi NLP while \citeauthor{Harish2020ACS} \shortcite{Harish2020ACS} offers an in-depth survey on regional Indic language processing. Developments in large pre-trained multilingual models like mBERT \cite{devlin2018bert}, XLM-RoBERTa \cite{conneau-etal-2020-unsupervised}, DistilmBERT \cite{Sanh2019DistilBERTAD}, IndicBERT \cite{kakwani2020indicnlpsuite} etc. have included Hindi and a variety of other regional Indic languages. However, as mentioned in the previous section, the lack of annotated data is a challenge that many in the Hindi NLP space are trying to overcome.

\section{Corpora}
The Automated Student Assessment (ASAP) corpus has benchmarked a variety of state-of-the-art English language models \cite{ke2019automated} ever since its release. Due to the ASAP corpus’ ubiquity in AES research and its depth in terms of volume and a mix of narrative, expository and source-dependent response prompts, we decided to focus our analysis on a Hindi-translated version of the ASAP dataset itself (ASAP-Hindi).
Translation has seen its way through the NLP discourse, especially in the development of multilingual corpora and models. The use of translated data, however, challenges the validity and quality of the corpus itself, with difficulty in determining the quality gain or loss in specific language-related attributes after translation is applied. We verified random subsets from the translated corpus with the help of both skilled bilingual speakers and expert academics. A common observation was that modern neural machine-translation engines have a proclivity to correct spelling mistakes, but the knowledge distilled and a majority of syntactic and semantic features are retained. Prompts 7 and 8 on the ASAP-Hindi dataset have a distinctly atypical scoring nature to the other prompts and real-world scoring rubrics. The first six prompts help generalize results to a range of AES contexts.

We also built our own Hindi language corpus consisting of 126 real-world essays written by students between 18-20 years old. The submissions were taken as an online essay writing competition, consisting of a single prompt with “The essence of Travel in one’s life” as the central theme. The average essay length is 224 words per essay. The writers of these essays are bilingual, with proficiency in both English and Hindi, with Hindi being their first language. Although much smaller in scale, the use of this corpus helps further corroborate our methods and gives a better understanding of how our methods work on real-world data. 

The scoring for the responses was in accordance with a comprehensive rubric that we prepared which took into account both subjectivity scores and attribute-based aspects. The subjectivity scores for each essay capture a general idea and take into account the evaluator’s general perception of the essay. Using an attribute-based approach, scores are provided for factors such as length sufficiency, coherence, relevance to prompt, argument quality, and vocabulary, which are more objective in nature. We combine these scores to obtain a final holistic score within a [0-12] range. 

This process is in accordance with a variety of previous scoring techniques in both English and AES in other languages. An expert panel of three Hindi academics performed the scoring following the rubric provided. Scorers 1-2, 1-3, and 2-3 have high inter-rater reliability \cite{cohen1968weighted} of 0.831, 0.798, and 0.867, respectively. These scores are comparatively higher than the ones calculated for the ASAP dataset. To further reduce the cognitive bias, we average the three scores to obtain the final score. Our final dataset contains responses from the eight prompts of the ASAP-Hindi corpus and our organic prompt.

\begin{table}[]
\fontsize{9}{9}\selectfont 
\setlength{\fboxsep}{0pt}%
  \setlength{\fboxrule}{1pt}%
  \fbox{%
{\renewcommand{\arraystretch}{1.5}%
\resizebox{\columnwidth}{!}{\begin{tabular}{|c|c|c|c|}
\hline
\multicolumn{1}{|l|}{}                                                                            & \textbf{Prompt} & \textbf{Essay Count} & \textbf{Score Range} \\ \hline
\multirow{8}{*}{\textbf{\begin{tabular}[c]{@{}c@{}}Hindi Translated \\ ASAP Corpus\end{tabular}}} & P1               & 1783                 & 2-12                 \\ \cline{2-4} 
                                                                                                  & P2               & 1800                 & 1-6                  \\ \cline{2-4} 
                                                                                                  & P3               & 1726                 & 0-3                  \\ \cline{2-4} 
                                                                                                  & P4               & 1772                 & 0-3                  \\ \cline{2-4} 
                                                                                                  & P5               & 1805                 & 0-4                  \\ \cline{2-4} 
                                                                                                  & P6               & 1800                 & 0-4
                                      \\ \cline{2-4} 
                                                                                                  & P7               & 1569                 & 0-30   
                                       \\ \cline{2-4} 
                                                                                                  & P8               & 723                 & 0-60                                                            
                                                                                                 \\ \hline
\textbf{Organic Corpus}     &     Travel                                                                                  & 126                  & 0-12                 \\ \hline
\end{tabular}}%
}%
}
\caption{Description of the prompts used for Hindi-AES}
\end{table}

\begin{figure}[h]
    \centering
    \includegraphics[scale=1]{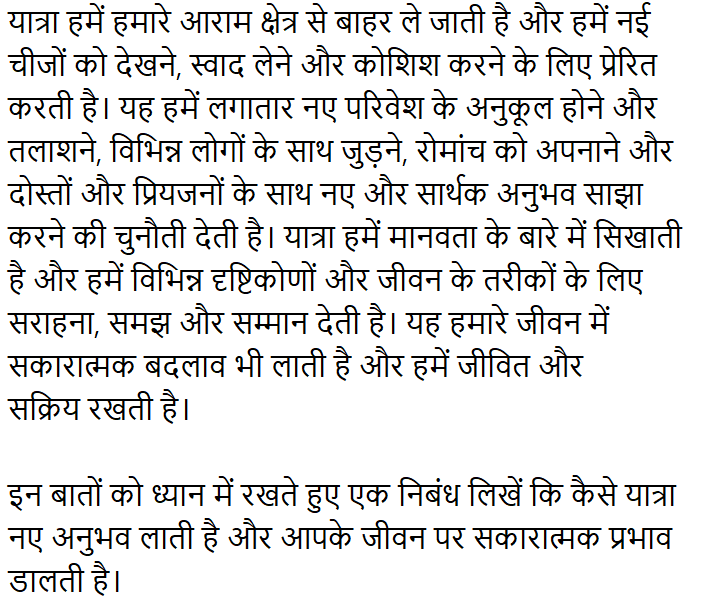}
    \caption{Prompt for the organic corpus}
\end{figure}

Figure 1 displays the prompt used for the organic corpus. Its translation in English is as follows: \emph{``Travel takes us out of our comfort zone and inspires us to see, taste and try new things. It challenges us to constantly adapt and explore new environments, connect with different people, embrace adventure, and share new and meaningful experiences with friends and loved ones. Travel teaches us about humanity and gives us an appreciation, understanding, and respect for different perspectives and ways of life. It also brings positive changes in our lives and keeps us alive and active. Keeping these things in mind, write an essay on how travel brings new experiences and positively impacts your life."}

\section{Methodology}
In our study we employ various methods ranging from classical machine learning approaches employing feature-extraction, to advanced fine-tuned transformer-based pre-trained models. We implement four models for classical regression and classification approaches; Linear Regression, Support Vector Regression (SVR), Random Forest and XGBoost. These leverage features like essay length, average sentence length, average word length, readability scores, semantic overlap, and vocabulary size. For neural network-based architecture, we implement a Bidirectional Long Short-Term Model (BiLSTM), a Convolutional Neural Network (CNN) Model, a CNN + LSTM coupled with an attention mechanism, and the popular SKIPFLOW model. Lastly, we fine-tune multiple pre-trained multilingual language models. A wide variety of approaches are used to effectively benchmark corresponding approaches from English AES research, in the Hindi language domain. 

\subsection{Feature Extraction}
Our features are largely inspired by the different dimensions of essay quality mentioned in \citeauthor{ke2019automated} \shortcite{ke2019automated}, readability metrics presented in \citeauthor{inproceedings} \shortcite{inproceedings} and the attributes presented in the ASAP++ dataset \cite{mathias-bhattacharyya-2018-asap}. Due to the tendency of neural translation engines to correct spelling, we do not explicitly employ a spelling-based feature in our feature selection. The extracted features are leveraged by our Linear Regression, SVR, Random Forest and XGBoost models. We extract and use the following six features:

\begin{itemize}
\item \textbf{Essay Length: }The essay's length is determined by the total number of words it contains.
\item \textbf{Average Sentence Length: }The average sentence length is calculated by adding up the lengths of all sentences and then dividing them by the number of sentences.
\item \textbf{Average Word Length: }The average word length is calculated by adding up the lengths of all words and then dividing them by the number of words.
\item \textbf{Readability Scores: }\citeauthor{inproceedings} \shortcite{inproceedings} presents this as a readability metric for Hindi and Bangla. Structural features such as Average Sentence Length (ASL), Average Word Length (AWL), Number of PolySyllabic Words (PSW), Number of Jukta-Akshars (JUK), and more are examined. They utilize Spearman’s rank correlation coefficient to analyze these structural features. Our readability scores are calculated using the following formula, based on the results of their regression analysis:

\[-2.34 + 2.14 * AWL + 0.01 * PSW\]

\item \textbf{Vocabulary and OOV words: }An essay’s score highly depends on its vocabulary size. Here in our experiments, we have simply taken into consideration the count of unique and out-of-vocabulary words (OOVs) in the essay to get the vocabulary size. Machine translation phonetically renders such OOVs into the target language (Hindi). Although if the OOV word is in frequent occurrence, we discard its use.

\item \textbf{Semantic Overlap and Coherence: }An essay’s score depends on how well it is connected and coherent. To determine this we have used mBERT and the mechanism used in SKIPFLOW for generating neural coherence features \cite{tay2018skipflow}. We calculated the semantic similarity of two sentences (using mBERT) at a distance of four sentences. As a result, we can capture the coherence and semantic overlap of the essay. Finally, after capturing the scores of the sentence pairs we averaged them out.

\end{itemize}

\subsection{Neural Approaches}
Deep neural networks like Convolutional Neural Networks (CNNs), Recurrent Neural Networks (RNNs), and Long Short-Term Memory networks (LSTMs) have facilitated the development of end-to-end machine learning workflows which do not rely on handcrafted features. We implement four popular approaches for AES using neural methods: BiLSTM, CNN, CNN + LSTM + Attention Mechanism, and SKIPFLOW. 
The training pipeline consists of an initial padding process for every sentence to equalize lengths, followed by a tokenization process using the IndicNLP Tokenizer. Word embeddings for the tokenized sentences were obtained using fastText(Wiki) pre-trained word vectors for Hindi \cite{bojanowski-etal-2017-enriching}. These word embeddings are passed to the following model architectures:

\begin{itemize}
\item \textbf{BiLSTM:} A Bidirectional LSTM, or BiLSTM, is a sequence processing model that consists of two LSTMs processing the sequence in both forward and backward directions. We use a BiLSTM over a normal LSTM as essays are dynamically and artistically written literary devices including references to both previous and forthcoming statements. Word embeddings are passed to a model with three BiLSTM layers and two dense layers to obtain the final score. We use Rectified Linear Unit (ReLu) activation functions between hidden layers and apply Batch Normalization after each BiLSTM layer. The first dense layer is followed by a Dropout layer.

\item \textbf{CNN:} When looked at from a different perspective, short-text responses and essays have many literary intricacies and relations that are more concentrated in short windows. Thus, using a CNN seems intuitive in capturing these short dependencies over a fixed window size. We use a CNN with three 1-Dimensional convolutional hidden layers separated by pooling layers, Swish \cite{ramachandran2017searching} non-linearities, and Batch Normalisation layers. 

\item \textbf{CNN + LSTM + Attention:} We use the model proposed by \citeauthor{dong-etal-2017-attention} \shortcite{dong-etal-2017-attention} to score the essays based on sentence representations using a model with CNN, LSTM, and Attention layers. Initially, a 1-Dimensional convolutional layer is used to extract features from a text in a way that is intuitively similar to window-based feature extraction. LSTM layers are used after that to process sequences. Finally, an attention layer is used to pool the outputs. 

\item \textbf{SKIPFLOW:} We implement the SKIPFLOW model introduced by \citeauthor{tay2018skipflow} \shortcite{tay2018skipflow}. SKIPFLOW proposes a novel method to calculate the textual coherence in the essays by modeling the relationships between snapshots of the hidden representations of a long short-term memory (LSTM) network as it reads. Subsequently, the semantic relationships between multiple snapshots are used as auxiliary features for prediction. 

\end{itemize}

\subsection{Fine-Tuned Transformers}
A transformer is a deep neural architecture that uses an attention mechanism to handle sequences of ordered data. Transformers include several layers each containing a multi-head self-attention network and a position-wise feed-forward network \cite{NIPS2017_3f5ee243}. Unlike LSTMs, Transformers use non-sequential processing, they do not process words in series but rather as a whole, allowing Transformers to take lesser time steps to process input in comparison to LSTMs and RNNs. Transformers also do not suffer from long-term dependency problems credit to their self-attention mechanism. 

Multilingual language models use BERT-based architecture and its variants and replace the corpus being used for the task of pre-training in English with the one containing text from multiple languages. Along with this, a few multilingual language models also introduce an additional pre-training task, for instance, Translation Language Modeling (TLM) in the case of XLM-R and MuRIL. IndicBERT and MuRIL are the two prominent Multilingual Language Models which specifically focus on regional Indic languages.	

We fine-tune prominent pre-trained multilingual transformer language models: Multilingual BERT (mBERT), DistilmBERT, XLM-Roberta, MuRIL, and IndicBERT. The tokenized prompt and essay are fed to the transformer model as input. The input is tokenized with a [CLS] token at the beginning. [SEP] tokens are added at the end of both the prompt and the essay to differentiate them. The prompt and the essay sequences are fed to the transformer encoder and become hidden layer sequences. These sequences are passed to a simple feed-forward network, consisting of two hidden layers, attached to the base transformer architecture and a final numerical score is obtained.

\section{Experiments}
This section describes the experimental procedure, evaluation metric, empirical results, and comparison with published results on prominent English AES models.

\begin{table*}[h!]
\fontsize{9}{9}\selectfont 
\begin{center}
\setlength{\fboxsep}{0pt}%
  \setlength{\fboxrule}{1pt}%
  \fbox{%
{\renewcommand{\arraystretch}{1.5}%
\begin{tabular}{|c|c|cccccccc|c|c|}
\hline \bf ID&\bf Hindi AES Models& \bf P1& \bf P2& \bf P3& \bf P4& \bf P5& \bf P6& \bf P7& \bf P8& \bf Average & \bf Organic\\ \hline

1& SVR                  & 0.799 & 0.612 & 0.605  & 0.657 & 0.797 & 0.630 & 0.400 & 0.380 & 0.610 & 0.579\\
2 & Linear Regression   & \bf{0.800} & 0.614 & 0.588  & 0.624 & 0.768 & 0.605 & 0.680 & 0.635 & 0.664 & 0.681 \\
3& Random Forest        & 0.705 & 0.608 & 0.621  & 0.685 & 0.791 & 0.665 & 0.652 & 0.560 & 0.661 & 0.762\\
4& XGBoost              & 0.794 & \bf{0.667} & 0.573  & 0.676 & 0.792 & 0.653 & 0.713 & \bf{0.641} & 0.688 & 0.827\\

\hline
5&CNN                    & 0.571 & 0.513 & 0.529 & 0.614 & 0.657 & 0.703 & 0.521 & 0.426 & 0.566 & 0.762\\
6&BiLSTM                 & 0.631 & 0.517 & 0.612 & 0.703 & 0.643 & 0.713 & 0.607 & 0.443 & 0.608 & 0.842 \\
7&CNN + LSTM + Attention & 0.723 & 0.597 & 0.677 & 0.711 & 0.781 & 0.791 & 0.701 & 0.593 & 0.696 & 0.827 \\
8&SKIPFLOW LSTM (Tensor) & 0.742 & 0.621 & 0.695 & 0.731 & 0.804 & 0.777 & 0.717 & 0.619 & 0.713 & 0.812\\

\hline
9 & mBERT       & 0.683 & 0.652 & \bf{0.711} & 0.775 & 0.828 & 0.785 & 0.781 & 0.548     & 0.720     & \bf{0.852}\\
10& DistilmBERT & 0.661 & 0.592 & 0.698 & 0.766 & 0.825 & 0.793 & 0.785 & 0.596     & 0.714     & 0.784\\
11& XLM-RoBERTa & 0.758 & 0.585 & 0.692 & \bf{0.809} & \bf{0.834} & \bf{0.822} & \bf{0.794} & 0.639  & 0.741$^*$    & 0.831\\

\hline
12& MuRIL       & 0.620 & 0.412 & 0.528 & 0.756 & 0.812 & 0.713 & 0.547 & 0.327  & 0.589 & 0.528\\
13& IndicBERT   & 0.651 & 0.489 & 0.659 & 0.751 & 0.799 & 0.784 & 0.708 & 0.412     & 0.656 & 0.796\\

\hline
\end{tabular}%
}%
}
\end{center}
\caption{\label{font-t2} Experiment results of all models in terms of QWK on ASAP-Hindi corpus and the organic prompt. The bold number is the best performance for each prompt. The best average QWK is annotated with $^*$.}
\end{table*}

\begin{table*}[h!]
\fontsize{9}{9}\selectfont 
\begin{center}
\setlength{\fboxsep}{0pt}%
  \setlength{\fboxrule}{1pt}%
  \fbox{%
{\renewcommand{\arraystretch}{1.5}%
\begin{tabular}{|c|c|cccccccc|c|c|}
\hline \bf ID&\bf English AES Models& \bf P1& \bf P2& \bf P3& \bf P4& \bf P5& \bf P6& \bf P7& \bf P8& \bf Average \bf \\ \hline
1&EASE (SVR)                    & 0.781  & 0.621 & 0.630 &0.749  & 0.782 &0.771 &0.727 &0.534  &0.699  \\
2&CNN + LSTM + Attention        & 0.822  & 0.682 & 0.672 &0.814  & 0.803 &0.811 &0.801 &0.705  &0.764  \\
3&SKIPFLOW LSTM(Tensor)         & \bf0.832  & 0.684 & 0.695 & 0.788 & 0.815 & 0.810& 0.800 &0.697 & 0.764 \\
4&R\textsuperscript{2}BERT      & 0.817  & \bf0.719 & \bf0.698 & \bf0.845 & \bf0.841 & \bf0.847& \bf0.839& \bf0.726& \bf0.791\\
\hline
\end{tabular}%
}%
}
\end{center}
\caption{\label{font-t2} Published results on prominent models for AES in English. All results in terms of QWK score on the original ASAP Corpus.}


\end{table*}

\subsection{Experimental Setup}
We conduct prompt-specific experimentation in accordance with \citeauthor{taghipour2016neural} \shortcite{taghipour2016neural}. While it might seem ideal to train prompts together, it's important to note that each prompt might include genres that are in sharp contrast to one another, such as narrative or argumentative essays. Prompts can also be scored differently according to the level of the students and the scoring rubrics. This makes training prompts together extremely challenging. 

For feature-based methods, the text was pre-processed to filter out stopwords, named-entities and mentions (denoted by ‘@’ symbols in the ASAP Dataset). The custom feature scores were normalized to improve the stability of the model. The neural LSTM and CNN models were all trained for 100 epochs with a learning rate of 1e-4. For fine-tuning the pre-trained multilingual transformer models we use the AdamW optimizer \cite{loshchilov2017decoupled}, which is a stochastic optimization method that modifies the typical implementation of weight decay in Adam \cite{kingma2014adam}, by decoupling weight decay from the gradient update. We set our learning rate to 5e-5 which decreases linearly to 0. We train all models on a 6-8 epoch range, depending on the base transformer model’s ability to learn. 

For our experiments on all prompts (including our organic prompt corpus) we use a 60/20/20 split for train, validation and test sets. Our normalization process keeps all score ranges within [0,1]. To calculate the Quadratic Weighted Kappa (QWK) scores, the scores are re-scaled to the original prompt-specific scale for prediction.

\subsection{Evaluation Metric}
We use the Quadratic Weighted Kappa (QWK), which is a common measure for evaluating and comparing AES methods \cite{cohen1968weighted}. A key reason for its prevalence is its particular sensitivity to differences in scores and its ability to take into account chance agreements. QWK score generally ranges from 0 to 1.  When the agreement is lower than expected by chance, the score becomes negative.
QWK Score is calculated as follows. Initially, a weight matrix W is created in accordance to Equation 1:

\begin{equation}
\textbf{W}{i,j} = \frac{(i-j)^2}{(N-1)^2} 
\end{equation}

If N is the total number of possible ratings, i and j represent the reference rating (given by a human annotator) and hypothesis rating (awarded by an AES system),  The outer product of the reference and hypotheses rating histogram vectors results in an expected count matrix E. This matrix is normalized so that the elements in E and O have the same sum. Finally, given the matrices O and E, the QWK score is calculated according to Equation 2:

\begin{equation}
\kappa = 1 - \frac{\sum{i,j}w_{i,j}O_{i,j}}{\sum_{i,j}w_{i,j}E_{i,j}}    
\end{equation}

\subsection{Results and Comparison}
In this section, we present the results of our experimentation. Table 2 (rows 1-13) reports the empirical results (QWK scores) on all our approaches on the ASAP Hindi Dataset and our organic prompt. We have also provided an average for results on the ASAP-Hindi set for a better comparison of the models across prompts. As a baseline for AES in Hindi does not exist, Table 3 (rows 1-4) presents published results on prominent models for AES in English. These results provide a benchmark upper limit for us to compare our scores, across all types of models. According to Tables 2 and 3, all 13 models implemented were able to learn the task and perform competitively when compared to results on the English models.

Results from the feature-based (Table 2, rows1-4)  approaches were also competitive, with Linear Regression and XGBoost outperforming all other models on prompt 1 and prompts 2 and 8 respectively. Averages for the feature-extraction models were not far off from the Ease SVR average benchmark in Table 3, with the lowest and highest averages falling 0.078 and 0.011 points short of the benchmark, respectively.

Although the neural models (Table 2, rows 5-8) did not outperform other models on any prompt, their averages were higher than the feature-based models and lower than the fine-tuned transformers, providing decent transitional results from classical to advanced end-to-end methods. SKIPFLOW (Table 2, row-8) gave the highest average score amongst these, followed closely by the CNN + LSTM + Attention average. The SKIPFLOW average fell 0.051 points short off it’s original implementation in English.

On average, the fine-tuned highly multilingual transformers (Table 2, rows 9-11) gave higher results across all prompts. The fine-tuned mBERT model gives the maximum score for prompt 3. The fine-tuned XLM-R model outperforms all other models on four of the eight ASAP-Hindi prompts (Prompts 4, 5, 6, 7), giving the maximum average QWK as well thereby, establishing a state-of-the-art for AES in Hindi. Compared to R\textsuperscript{2}BERT’s average the fine-tuned XLM-R is 0.050 points short. Results on the mBERT model closely follow the results on the XLM-R model. The fine-tuned Indic transformers (Table 2, rows 12 and 13) did not perform comparably well. While IndicBERT did try to compete and learn, MuRIL's behavior during the training process was highly inconsistent, resulting in unpredictable results. This behavior could be attributed to a variety of factors which will be discussed in the next sub-section.

Results on the organic prompt were favourable (in comparison to results on the ASAP-Hindi set) with the fine-tuned mBERT model giving the highest QWK score. It is important to note that in contrast to the other prompts, the QWK scores obtained for organic prompt showed a slightly higher variance during training. It is likely that this is a result of the organic set's smaller magnitude compared to the ASAP-Hindi Dataset.

\subsection{Analysis and Discussion}
A general observation that is consistent with both previously established English AES results and our study, is that results on prompts 4,5, and 6 are higher than the other prompts for the ASAP dataset. Prompts requiring source-dependent responses perform better during training as compared to narrative, persuasive or expository prompts possibly due to a general consistency in syntax, coherence, and availability of source material. Such prompts are coupled with more balanced real-world rubrics allowing for consistency all throughout the writing and the scoring process, making such prompts ideal for generalization. In contrast generalization of ideas is slightly more difficult on prompts that allow for persuasive and expository discussions due to variance in human thought and cognition. Prompts with rubrics that are not consistent with real-world rubrics such as prompt-8, give results much worse than the aforementioned source-dependent prompts. 

The more discrete nature of feature-based approaches allows for consistent performance, but their failure to understand nuance and interpret content limits their potential for improvement when compared to large pre-trained language models. Given the considerably larger scale (parameters and pre-training data) of XLM-R and mBERT they significantly outperform the other models. Although DistilmBERT is approximately 40\% lighter than mBERT, it performs almost equally well and was the fastest to train among all large language models.

A result observed was the relatively poor performance of the Indic language models, especially MuRIL. Several plausible explanations might explain this finding. IndicBERT’s performance can be attributed to its base architecture being an ALBERT model which is almost 90\% lighter than the BERT-base. IndicBERT may have been disadvantaged by the lack of trainable parameters because of this, resulting in it not competing with mBERT and XLM-R. MuRIL, however, does have the parameter size to compete. \citeauthor{patil-etal-2021-vyakarana} \shortcite{patil-etal-2021-vyakarana} provides compelling evidence as to why MuRIL might fail to perform on syntactically-complex tasks such as AES, it includes the following points: Both IndicBERT and MuRIL perform masked word-level language modeling and do not have a sentence level pre-training task. An important aspect of AES is the morpho-syntactical relationship between ideas and sentences. mBERT, XLM-R, and DistilmBERT are highly multilingual language models pre-trained on more than 100 languages. It may provide them with linguistic and typological generalizations needed to model morpho-syntax more effectively than Indic models, which are trained on a small number of Indic languages and English. Another factor to add to this is that monolingual and multilingual large language models show syntactic localization across their layers which makes them perform better at complex tasks with long-range syntactical dependencies. In comparison, Indic language models IndicBERT and MuRIL show little localization, with MuRIL showing the least localization across all layers amongst all models. The syntactic directness of source-dependent responses is possibly why MuRIL still remains competitive on prompts 4, 5, and 6.

\section{Conclusion and Future Work}

In this study, we implement and analyze various methods for AES in Hindi ranging from classical feature-based Machine Learning (ML) to advanced end-to-end models to set benchmarks and the state-of-the-art for AES in Hindi. We also introduce a single-prompt corpus of student-written essays in Hindi to further substantiate our findings from the Hindi-Translated ASAP corpus. The results of our experiments shed new light on AES research. We obtain competitive results when compared to the benchmark and the state-of-the-art methods in English AES. We also try to explain and analyze the results obtained using our models, attributing to the idiosyncrasies of the models as well as the nature of the prompts on which they are tested. 

In view of the fact that AES in Hindi is particularly unexplored, multiple future research directions are possible. We plan to extend our work by scaling the organic corpus (both in terms of essays per prompt and the type of prompts) and proposing architectures that learn the syntactical features and nuances of Hindi by leveraging the trait-based scores that are included with the corpus. It is reasonable to use a Hindi-translated version of the ASAP dataset as precedent, but a comprehensive Hindi corpus is essential for AES in Hindi. A larger corpus facilitates more nuanced learning, which enables models to generalize from a wider spectrum of results. 

We obtain the most favorable results on our fine-tuned pre-trained multilingual transformer language models and to push these results further, we plan to try different training optimization methods including domain adversarial training, multi-scale essay representation approaches and more. Using such training optimization methods might improve performance on Hindi which is more morpho-syntactically complex than English. For the same reason, introducing linguistic knowledge to segment at a more reasonable scale may bring further improvement. We also hope to push the results on the Indic language models IndicBERT and MuRIL through these optimization strategies. Due to the prominence of regional languages in Indian communities, a multilingual essay evaluation approach will allow for more diversity in essay writing and large-scale examinations.

\section*{Acknowledgements}
Rajiv Ratn Shah was partly supported by the Infosys Center for Artificial Intelligence and the Center of Design and New Media at IIIT Delhi, India.

\fontsize{9.4pt}{10.4pt} \selectfont
\bibliography{aaai23}
\end{document}